\documentclass[runningheads]{llncs}
\usepackage[T1]{fontenc}
\usepackage{newtxtext}
\usepackage[varvw]{newtxmath}
\usepackage{array}
\usepackage{amsmath}
\usepackage{graphicx}
\usepackage{adjustbox}
\usepackage{float}
\usepackage{textcomp}
\usepackage{xcolor}
\usepackage{multirow}
\usepackage{placeins}
\newcommand{\necatablestyle}{%
    \small
    \setlength{\tabcolsep}{2pt}%
    \renewcommand{\arraystretch}{1.05}%
}

\begin{document}

\title{NECA: Neighbor Event Curve Augmentation for Multi-Horizon ICU Deterioration Prediction}

\titlerunning{NECA for ICU Deterioration Prediction}
\author{
Chenyang Wang\textsuperscript{1},
Theresa Jiawei Tang\textsuperscript{1},
Xinran Liu\textsuperscript{1},\\
Xinyu Lu\textsuperscript{2},
Xulei Yang\textsuperscript{3}\textsuperscript{*}
}
\authorrunning{Chenyang Wang}
\institute{\textsuperscript{1}\textit{Department of Statistics and Data Science,
National University of Singapore, Singapore}\\
\textsuperscript{2}\textit{School of Computing,
National University of Singapore, Singapore}\\
\textsuperscript{3}\textit{Institute for Infocomm Research (I$^2$R),
A*STAR, Singapore}\\
Contact: e1155460@u.nus.edu
}
\maketitle

\begin{abstract}
Early prediction of ICU organ dysfunction events supports timely clinical monitoring and intervention, but estimating short-term organ-specific risk from early ICU records remains challenging due to irregular measurements, heterogeneous clinical variables, and the need to predict risk across multiple future horizons. Existing supervised models usually treat each patient as an isolated sample and do not explicitly use the observed event timing of similar historical patients. We propose Neighbor Event-Curve Augmentation (NECA), a model-agnostic method that retrieves training-set patients with similar early-window profiles and converts their subsequent organ dysfunction event onset times into cumulative event-curve features over 6h, 12h, 24h, and 42h. These features are concatenated with the original patient representation and used by standard downstream predictors. On PhysioNet 2012 and a 4,000-stay MIMIC-IV subset, NECA improves average PR-AUC and ROC-AUC across four organ dysfunction event tasks and seven supervised learners. The largest gain appears for neurological dysfunction events, where task-level average PR-AUC increases from 0.461 to 0.557 on PhysioNet 2012. Ablations show that clinically meaningful retrieval and horizon-specific neighbor risk drive most of the improvement, while the full cumulative curve adds task-dependent temporal context. These results support similar-patient event timing as a structured temporal risk signal for ICU organ dysfunction event prediction. The source code will be publicly released upon acceptance of this paper.
\end{abstract}

\section{Introduction}
Early recognition of organ dysfunction in intensive care units (ICUs) is a central problem in biomedical and health informatics. Machine-learning early warning systems, such as TREWS for sepsis, show that timely risk identification can support earlier intervention and improve outcomes \cite{adams2022trews}. ICU patients are monitored through electronic health records (EHRs), irregular physiological measurements, laboratory tests, and clinical interventions, producing rich but heterogeneous time-series data. However, many bedside decisions are not only about whether a patient is high risk, but also about which organ dysfunction event is likely to be observed next and within what time scale \cite{blythe2024timeevent,do2025multihorizon}. In this work, EHR-observed organ dysfunction events are the first newly observed post-window task-specific clinical abnormalities, such as renal dysfunction, neurological dysfunction, electrolyte/metabolic abnormality, or inflammatory abnormality, according to predefined EHR measurement criteria.

Existing ICU prediction methods usually learn a supervised mapping from the current patient record to a target outcome. Logistic regression, gradient-boosted trees, TabPFN, and temporal models such as LSTM can estimate future risk from early-window features or irregular clinical time series \cite{friedman2001greedy,hosmer2013applied,chen2016xgboost,ke2017lightgbm,hollmann2025tabpfn,hochreiter1997lstm}. Stronger temporal encoders can better model sparsity and temporal dependence, but they still usually produce a risk score for a fixed horizon. They do not explicitly use the subsequent trajectories of clinically similar historical patients, and therefore provide limited case-based evidence about whether organ dysfunction events tend to occur within 6, 12, 24, or 42 hours.

Our motivation is to bring case-based temporal evidence into early ICU prediction. Patient-similarity work shows that clinically similar historical patients can inform personalized risk estimation \cite{ng2015personalized,sharafoddini2017patientsimilarity}. If similar ICU patients later showed renal dysfunction, neurological dysfunction, inflammatory abnormality, or electrolyte/metabolic abnormality shortly after the early window, their trajectories provide local temporal evidence for the current patient. Instead of only using similar patients as static neighbors or graph nodes, we model when their post-window events appeared. This converts neighbor information from a static risk estimate into a multi-horizon event trajectory that can be used by standard predictors.

We propose Neighbor Event-Curve Augmentation (NECA), a model-agnostic method for early-window multi-horizon organ dysfunction event prediction. For each query patient, NECA retrieves similar training-set patients using only early-window information and summarizes their future event onset times as cumulative event-curve features over 6h, 12h, 24h, and 42h. These features are concatenated with the original patient representation without changing downstream architectures. NECA is temporally valid and interpretable: future information comes only from training-set neighbors, and each augmented feature is the fraction of similar patients who reached the event definition within a future horizon. Because each dimension corresponds to a concrete future horizon, the augmentation can be inspected as a local event-risk summary rather than an opaque latent embedding.

The main contributions are:

\begin{itemize}
    \item We formulate early ICU organ dysfunction event prediction as a multi-task, multi-horizon problem over four EHR-defined event categories within 6h, 12h, 24h, and 42h after an early window.
    \item We propose NECA, which transforms post-window event onset times of similar training-set patients into cumulative event-curve features used as retrieval-derived temporal risk signals.
    \item We evaluate four organ dysfunction event tasks and seven supervised learners, showing that event-curve features consistently complement supervised predictors, especially for neurological dysfunction events.
\end{itemize}

\section{Related Work}

\subsection{Early ICU Organ Dysfunction Event Prediction}

Early ICU prediction has been widely studied using physiological time-series and EHR data. Harutyunyan et al.~\cite{harutyunyan2019multitask} introduced a Medical Information Mart for Intensive Care III (MIMIC-III) benchmark for mortality, decompensation, length of stay, and phenotype classification. Chen et al.~\cite{chen2018dynamic} developed multi-task recurrent models for illness severity prediction. ConCare learns personalized clinical feature embeddings with feature-specific temporal encoders and healthcare-context attention~\cite{concare2020}. APRICOT-M uses state-space modeling for real-time ICU acuity, stability, mortality, discharge, and therapy prediction~\cite{contreras2025realtime}.

Beyond global acuity and mortality prediction, recent studies have considered organ-level or multi-label organ dysfunction prediction. SeqSNR models six ICU organ dysfunction events using shared multitask representations~\cite{roy2021multitask}, while IOC-MT uses graph attention for inter-organ correlations~\cite{zeng2025interorgan}. Other work studies early dynamic organ-system risk~\cite{soley2025early} and multi-label care escalation triggers including hemodynamic instability, renal compromise, and neurological events~\cite{bukhari2025early}. These studies motivate prediction at short horizons for specific organ systems. NECA additionally supplies each predictor with recorded outcomes from retrieved historical records.

\subsection{Retrieval-Augmented and Patient-Similarity Clinical Prediction}

Retrieval and knowledge augmentation have also been applied to EHR prediction. MedRetriever retrieves unstructured text for interpretable health risk prediction~\cite{medretriever}, while RAM-EHR uses dense retrieval over a multi-source clinical corpus to augment local EHR models~\cite{ramehr}. GraphCare and KARE incorporate biomedical knowledge graphs or retrieved community summaries~\cite{graphcare,kare}, and CLEAR demonstrates targeted retrieval for clinical information extraction~\cite{clear}. These studies motivate retrieval as a prediction component, although their evidence units and downstream tasks differ from NECA's historical ICU records.

Patient-similarity methods provide cross-patient evidence. Models using related trajectories learn patient representations~\cite{similarityaware}, while similarity graphs support ICU mortality and criticalness prediction~\cite{sbscgm}. NECA passes four cumulative neighbor event rates to the predictor, making this historical evidence inspectable. Prior work also studies patient retrieval and nearest-neighbor prediction.

\section{Methods}

\begin{figure}[t]
    \centering
    \includegraphics[width=\textwidth]{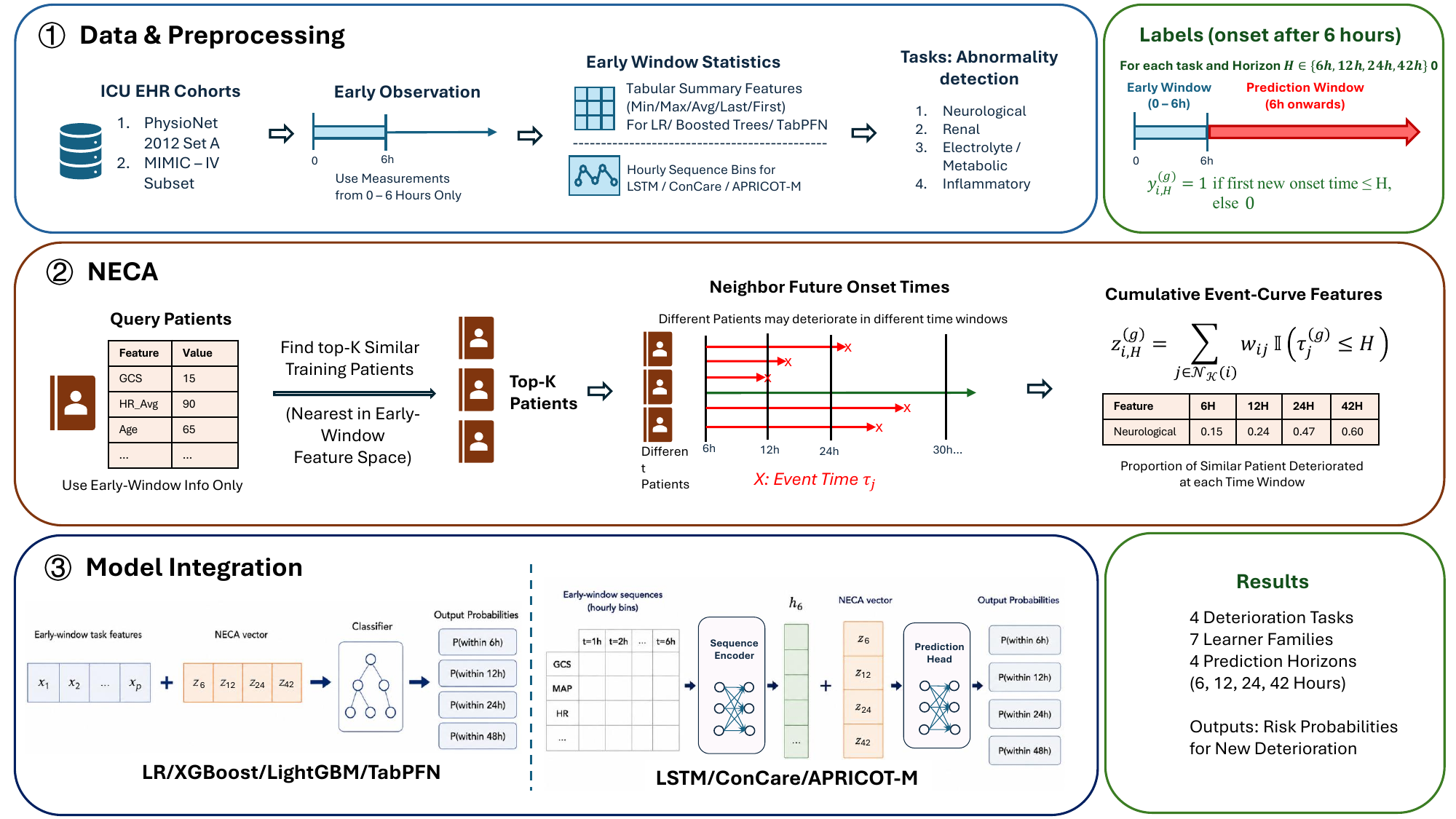}
    \caption{Overview of the proposed NECA framework for early ICU organ dysfunction event prediction. 
    The framework consists of data preprocessing, neighbor event-curve augmentation, model integration, and multi-horizon output prediction.}
    \label{fig:neca_framework}
\end{figure}

Figure~\ref{fig:neca_framework} shows the four steps of NECA. It constructs early-window representations, retrieves similar training records, summarizes their later event times, and adds the resulting event curve to a predictor.

\subsection{Data Sources and Cohort Construction}

We evaluate NECA on two ICU time-series datasets. PhysioNet/Computing in Cardiology Challenge 2012 Set A contains 4,000 ICU records with irregular physiological and laboratory measurements \cite{silva2012physionet} (Table~\ref{tab:data_summary}). We also use a subset of 4,000 ICU stays from MIMIC-IV \cite{johnson2023mimiciv}. Each record represents one ICU stay. The MIMIC-IV loader retains stays of adults aged at least 18 years, with a recorded ICU length of stay of at least 12 hours, from admissions containing exactly one ICU stay. It sorts the stays by identifier and samples 4,000 without replacement using a fixed random seed. A subject is an individual patient and can have records from multiple admissions. We represent a record as
\begin{equation}
    \mathcal{R}_i = \bigl(s_i,\{(v,t_{ij}^{(v)},x_{ij}^{(v)}):v\in\mathcal{V},\;j=1,\ldots,n_{iv}\}\bigr),
\end{equation}
Here $s_i$ denotes static information for record $i$, $\mathcal{V}$ is the set of clinical variables, and $n_{iv}$ is the number of observations of variable $v$ in that record. The value $x_{ij}^{(v)}$ was recorded at time $t_{ij}^{(v)}$ after ICU admission.

In the tables, HR means heart rate, RR means respiratory rate, MAP means mean arterial pressure, and SBP means systolic blood pressure. SpO$_2$ is peripheral oxygen saturation, PaO$_2$ is arterial oxygen partial pressure, and FiO$_2$ is the fraction of inspired oxygen. GCS means Glasgow Coma Scale, BUN means blood urea nitrogen, WBC means white blood cell count, and HCO$_3$ means bicarbonate. Cr and Temp denote creatinine and temperature.

\begin{table}[t]
\centering
\caption{Summary of ICU datasets and selected clinical variables.}
\label{tab:data_summary}
\necatablestyle
\begin{adjustbox}{max width=\columnwidth}
\begin{tabular}{p{0.2\linewidth} p{0.36\linewidth} p{0.32\linewidth}}
\hline
\textbf{Aspect} & \textbf{PhysioNet 2012 Set A} & \textbf{MIMIC-IV subset} \\
\hline
Cohort size & 4,000 ICU records & 4,000 ICU stays \\
Time origin & ICU admission & ICU admission \\
Sampling pattern & Sparse and irregular & Sparse and irregular \\
Static variables & Age, gender, height, weight, ICU type & Age, gender and mapped ICU variables \\
High-coverage variables & HR, temperature, GCS, creatinine, BUN, WBC, bicarbonate, urine output & HR, oxygen saturation, GCS, respiratory rate, creatinine, BUN, WBC, bicarbonate, platelets \\
Moderate/sparse variables & Lactate, pH, PaO$_2$, FiO$_2$, mechanical ventilation, bilirubin & Urine output, pH, lactate, FiO$_2$, mechanical ventilation \\
\hline
\end{tabular}
\end{adjustbox}
\end{table}

The early observation window spans $T_e=6$ hours after ICU admission. Measurements within $[0,T_e]$ provide prediction features. Later measurements determine event times and labels.

\subsection{Multi-Horizon Organ Dysfunction Event Prediction}

Let $\mathcal{G}$ contain the four endpoints introduced above. For endpoint $g\in\mathcal{G}$, the rule $A_g(\cdot)$ equals one when the measurements meet that endpoint's abnormality threshold. The notation $\mathcal{R}_i[0,t]$ denotes record $i$ through time $t$. We define $\tau_i^{(g)}$ as the time from the end of the early window to the first recorded event for endpoint $g$.
\begin{equation}
    \tau_i^{(g)} =
    \min \{t - T_e: A_g(\mathcal{R}_i[0,t]) = 1, \ t > T_e\}.
\end{equation}
If the record meets the event rule during the early window, we exclude it for that endpoint. The resulting prediction concerns newly recorded events. If relevant early measurements are absent, the rule evaluates false and the record can remain eligible. Early absence does not establish normal organ status. The labeler assigns $\tau_i^{(g)}=\infty$ when the available record contains no qualifying event. This also occurs when relevant follow-up measurements are missing or a stay ends before the prediction horizon. An infinite value therefore does not establish that the patient remained event-free.

For a prediction horizon $H \in \{6,12,24,42\}$ hours, the binary label is defined as
\begin{equation}
    y_{i,H}^{(g)} =
    \mathbb{I}(\tau_i^{(g)} \leq H).
\end{equation}
The final horizon is 42 hours after the early window, reaching 48 hours after ICU admission. The label $y_{i,H}^{(g)}=1$ means that record $i$ has a recorded event for endpoint $g$ within $H$ hours after the early window. We train a separate binary classifier for each endpoint and horizon. The result tables report eligible record counts.

The rules draw on the Sequential Organ Failure Assessment (SOFA), Sepsis-3, National Early Warning Score 2 (NEWS2), Kidney Disease Improving Global Outcomes (KDIGO) guidelines, and systemic inflammatory response syndrome (SIRS) criteria \cite{vincent1996sofa,singer2016sepsis3,rcp2017news2,welch2022news2,kdigo2012aki,bone1992sirs}. They do not implement any of these complete definitions.

The renal endpoint uses creatinine and BUN. The neurological endpoint uses GCS. The electrolyte/metabolic endpoint uses pH, bicarbonate, and lactate. The inflammatory endpoint uses WBC and temperature.

Measurement frequency, treatment, and documentation practice can affect whether an event is recorded. Our target is therefore the first newly recorded event that meets an endpoint rule after the early window. Table~\ref{tab:tasks} gives the rules and early-window variables.

\begin{table}[t]
\centering
\caption{EHR-observed endpoint rules and selected first-six-hour predictor variables.}
\label{tab:tasks}
\small\setlength{\tabcolsep}{3pt}\renewcommand{\arraystretch}{1.08}
\begin{tabular}{@{}>{\raggedright\arraybackslash}p{0.21\textwidth}>{\raggedright\arraybackslash}p{0.30\textwidth}>{\raggedright\arraybackslash}p{0.39\textwidth}@{}}
\hline
Endpoint & Recorded event rule & Early-window variables \\
\hline
Renal & Maximum Cr $>1.5$ mg/dL or BUN $>30$ mg/dL & Age, gender, minimum MAP, maximum Cr and BUN, last urine output, maximum lactate \\
Neurological & Last GCS $\leq12$ & Age, gender, minimum and last GCS, minimum MAP, maximum HR \\
Electrolyte / metabolic & Minimum pH $<7.35$, minimum HCO$_3$ $<22$, or maximum lactate $>2.0$ & Age, gender, maximum lactate, minimum pH and HCO$_3$, maximum Cr and BUN \\
Inflammatory & Maximum WBC $>12$ or temperature $>38^\circ$C & Age, gender, maximum WBC, maximum temperature, minimum platelets, maximum lactate \\
\hline
\end{tabular}
\end{table} 

\subsection{Early-Window Feature Representation}

For tabular models, each record becomes an endpoint-specific early-window feature vector
\begin{equation}
    \mathbf{x}_i^{(g)} \in \mathbb{R}^{d_g},
\end{equation}
where $d_g$ is the number of variables selected for endpoint $g$. The vector contains summary statistics from $[0,T_e]$, including first, last, minimum, and maximum values. Missing values are retained and later filled using medians computed from the training set.

For sequence models, we construct an hourly representation from the same early-window time series. The interval $[0,T_e]$ is divided into $T=\lceil T_e \rceil$ hourly bins, and observations within each bin are aggregated by taking the last value, yielding
\begin{equation}
    \mathbf{S}_i \in \mathbb{R}^{T \times F},
\end{equation}
where $F$ is the number of physiological and laboratory channels. LSTM, ConCare, and an APRICOT-M-style state-space model (SSM) use this tensor, so sequence baselines receive the same early-window information in temporally ordered form.

For LSTM, missing values are forward-filled and remaining entries are filled from training-set empirical values before channel-wise standardization. ConCare reorganizes the same tensor into feature-specific temporal streams before healthcare-context attention combines variable-level representations. The adapted state-space model uses the hourly tensor. All sequence-model preprocessing statistics are fitted on the training split only.

\subsection{Neighbor Event-Curve Augmentation}

NECA augments each early-window representation with recorded event times from retrieved neighbors. For query record $i$, it retrieves up to $K$ training records using only early-window information. We denote this set by $\mathcal{N}_K(i)$.

For each neighbor $j \in \mathcal{N}_K(i)$, the training data provide its recorded event time $\tau_j^{(g)}$. NECA summarizes these times at each prediction horizon $H$ using
\begin{equation}
    z_{i,H}^{(g)} =
    \sum_{j \in \mathcal{N}_K(i)}
    w_{ij}
    \mathbb{I}(\tau_j^{(g)} \leq H \ ),
\end{equation}
Here $w_{ij}$ is a neighbor weight that sums to one over $\mathcal{N}_K(i)$. In the main experiment, all retrieved neighbors receive equal weight. If exactly $K$ neighbors are retrieved, $w_{ij}=1/K$. A distance-weighted variant gives closer neighbors more weight. We collect the four rates for endpoint $g$ in the vector
\begin{equation}
    \mathbf{z}_i^{(g)}
    =
    \left[
    z_{i,6}^{(g)},
    z_{i,12}^{(g)},
    z_{i,24}^{(g)},
    z_{i,42}^{(g)}
    \right].
\end{equation}
Each entry is the weighted fraction of retrieved records with an event by its prediction horizon. The entries are nondecreasing because the horizons are cumulative. We call this vector the neighbor event curve.

For tabular models, we append the neighbor event curve to the early-window vector.
\begin{equation}
    \tilde{\mathbf{x}}_i^{(g)}
    =
    \left[
    \mathbf{x}_i^{(g)} ;
    \mathbf{z}_i^{(g)}
    \right].
\end{equation}

For sequence models, we append the neighbor event curve to the encoder's final hidden representation before prediction. We call a model trained without the event curve Raw and the corresponding augmented model Curve-NECA.

\subsection{Neighbor Retrieval and Leakage Control}

NECA can retrieve neighbors using different early-window features. In the reported main experiment, retrieval uses 27 standardized abnormality indicators, Euclidean distance, equal neighbor weights, and $K=100$. This value of $K$ was fixed for the reported results. Table~\ref{tab:retrieval_feature_details} lists the alternative retrieval feature sets used in ablations. Appendix~\ref{app:flag_definitions} defines the 27 indicators.

\begin{table}[t]
\centering
\caption{Retrieval feature presets used by NECA.}
\label{tab:retrieval_feature_details}
\necatablestyle
\begin{adjustbox}{max width=\columnwidth}
\begin{tabular}{@{}p{0.22\columnwidth} c p{0.58\columnwidth}@{}}
\hline
\textbf{Preset} & \textbf{Dim.} & \textbf{Description} \\
\hline
Endpoint-specific & 6--14 & Early-window variables selected for the target endpoint. \\
Demographic & 2 & Age and gender. \\
Physiological & 15 & Vital-sign and physiological summaries, including SBP, MAP, HR, RR, and SpO$_2$. \\
Laboratory & 17 & Laboratory and organ-status summaries, including PaO$_2$, FiO$_2$, temperature, GCS, WBC, Cr, BUN, urine output, lactate, pH, HCO$_3$, platelets, and bilirubin. \\
Abnormality indicators & 27 & Binary and normalized indicators covering hemodynamic, respiratory, renal, inflammatory, hepatic, coagulation, shock-like, multi-system, monitoring, and ICU-type signals. \\
\hline
\end{tabular}
\end{adjustbox}
\end{table}

For each random seed and endpoint, we split eligible records approximately 70\% for training, 10\% for validation, and 20\% for testing. When group counts permit, we stratify by recorded event time using 0--6h, 6--12h, 12--24h, 24--48h, and $>$48h or no recorded event. The split uses records as units, including in MIMIC-IV. Records from different admissions of one subject can therefore appear in different partitions. Imputation and scaling use only the training partition. Validation and test event curves use outcomes only from training neighbors. For a training query, we exclude its own record from retrieval. Other records from the same subject can remain. Training event curves are not cross-fitted, which would require each training query to retrieve only from other training folds. These choices can make performance optimistic. A subject-disjoint split and cross-fitted training event curves are needed to assess that risk.

\subsection{Prediction Models}

NECA can augment both tabular and sequence predictors. We evaluate seven supervised learners, including logistic regression, XGBoost, LightGBM, TabPFN, LSTM, ConCare, and an APRICOT-M-style SSM. We train a separate binary classifier for each endpoint $g$ and prediction horizon $H$.
\begin{equation}
    \hat{p}_{i,H}^{(g)}
    =
    f_{\theta}^{(g,H)}(\tilde{\mathbf{x}}_i^{(g)}),
\end{equation}
Here $f_{\theta}^{(g,H)}$ is a classifier with learned parameters $\theta$, and $\hat{p}_{i,H}^{(g)}$ is its predicted event probability for endpoint $g$ by horizon $H$.

For LSTM, ConCare, and the APRICOT-M-style SSM, we append the neighbor event curve after the sequence encoder. The resulting prediction is
\begin{equation}
    \hat{p}_{i,H}^{(g)}
    =
    \sigma
    \left(
    \mathbf{w}^{\top}
    [
    \mathbf{h}_i ;
    \mathbf{z}_i^{(g)}
    ]
    + b
    \right),
\end{equation}
Here $\mathbf{h}_i$ is the sequence encoder's final hidden representation, $\mathbf{z}_i^{(g)}$ is the neighbor event curve, $\mathbf{w}$ and $b$ are output-layer parameters, and $\sigma(\cdot)$ is the sigmoid function. Raw sequence models predict from $\mathbf{h}_i$. Curve-NECA predicts from $[\mathbf{h}_i;\mathbf{z}_i^{(g)}]$.

ConCare is an attention-based sequence baseline for irregular clinical time series. It learns personalized feature embeddings from feature-specific temporal streams and uses healthcare-context attention to combine clinical variables and static context \cite{concare2020}. We train ConCare separately for each endpoint and horizon using the same early-window tensor and labels as LSTM. Curve-NECA appends the four-dimensional neighbor event curve before ConCare's final classifier.

The APRICOT-M-style SSM is inspired by a model for real-time ICU acuity and therapy prediction \cite{contreras2025realtime}. We adapt its temporal encoder to each endpoint and horizon. This implementation does not reproduce the original APRICOT-M multi-task system. Its Raw version predicts from the early-window sequence. Curve-NECA appends the neighbor event curve to its hidden representation, as with LSTM and ConCare.

Model settings are fixed across Raw and Curve-NECA comparisons. Logistic regression uses balanced classes and $C=1$. XGBoost uses depth 5, 900 trees, and learning rate 0.05. LightGBM uses 63 leaves, 1500 trees, and learning rate 0.03. Sequence models use Adam, at most 40 epochs, and early stopping after seven epochs. TabPFN uses its pretrained classifier without tuning.

\subsection{Evaluation Protocol}

We evaluate area under the receiver operating characteristic curve (ROC-AUC) and area under the precision-recall curve (PR-AUC) on validation and test partitions. PR-AUC is particularly relevant because several endpoints are imbalanced (Table~\ref{tab:positive_rates}). Raw and Curve-NECA use the same split for each endpoint, learner, horizon, and seed. Each horizon cell reports mean test performance across seeds. We average across horizons, the seven supervised learners, and seeds to obtain each endpoint-level result.  The reported tables do not include paired confidence intervals, so small differences remain uncertain. In tables, PN12 denotes PhysioNet 2012 and MIMIC denotes the MIMIC-IV subset.

A k-nearest-neighbor (KNN) predictor uses the event rate among the same retrieved training neighbors as a direct reference. On PN12, its mean PR-AUC across horizons is 0.333, 0.355, 0.303, and 0.467 for neurological, renal, electrolyte/metabolic, and inflammatory endpoints, respectively. The tabular Raw models receive a small endpoint-specific feature set, while retrieval uses 27 cross-system indicators. The current comparison therefore cannot isolate the effect of neighbor outcomes from the additional input information. An information-matched comparison would evaluate the Raw features with all 27 indicators, the indicators alone, and the same combined inputs with NECA. The reported experiments do not include these controls.

\begin{table}[t]
\centering
\caption{Positive rates (\%) of EHR-observed events across prediction horizons.}
\label{tab:positive_rates}
\necatablestyle
\begin{adjustbox}{max width=\columnwidth}
\begin{tabular}{@{}llccccc c@{}}
\hline
\textbf{Data} & \textbf{Endpoint} & \textbf{N} 
& \textbf{6h} & \textbf{12h} 
& \textbf{24h} & \textbf{42h} 
& \textbf{Avg.} \\
\hline
\multirow{4}{*}{PN12} 
& Neuro. & 2040 & 15.1 & 20.6 & 28.6 & 33.7 & 24.5 \\
& Renal & 3171 & 10.1 & 20.8 & 27.3 & 37.3 & 23.9 \\
& Electrolyte & 2071 & 15.8 & 23.3 & 30.9 & 38.2 & 27.1 \\
& Inflam. & 2486 & 21.4 & 36.2 & 47.8 & 57.8 & 40.8 \\
\hline
\multirow{4}{*}{MIMIC} 
& Neuro. & 2650 & 7.3 & 9.4 & 12.2 & 13.9 & 10.7 \\
& Renal & 3401 & 10.5 & 17.1 & 23.0 & 27.4 & 19.5 \\
& Electrolyte & 2743 & 15.3 & 21.8 & 29.2 & 33.3 & 24.9 \\
& Inflam. & 1989 & 10.1 & 15.9 & 20.8 & 24.5 & 17.8 \\
\hline
\end{tabular}
\end{adjustbox}
\end{table}

\section{Results}

\subsection{Overall Performance of NECA}

\begin{table}[t]
\centering
\caption{Mean test performance across four horizons and seven supervised learners. PN12 is PhysioNet 2012; MIMIC is the 4,000-stay MIMIC-IV subset.}
\label{tab:physionet_task_avg}\label{tab:mimic_task_avg}
\small\setlength{\tabcolsep}{4pt}
\begin{tabular}{llcccc}
\hline
 & & \multicolumn{2}{c}{PR-AUC} & \multicolumn{2}{c}{ROC-AUC} \\
\cline{3-6}
Dataset & Endpoint & Raw & Curve-NECA & Raw & Curve-NECA \\
\hline
PN12 & Neurological & 0.461 & 0.557 & 0.689 & 0.766 \\
 & Renal & 0.353 & 0.387 & 0.711 & 0.740 \\
 & Electrolyte/metabolic & 0.357 & 0.378 & 0.671 & 0.692 \\
 & Inflammatory & 0.514 & 0.538 & 0.692 & 0.727 \\
\hline
MIMIC & Neurological & 0.419 & 0.520 & 0.749 & 0.809 \\
 & Renal & 0.319 & 0.345 & 0.668 & 0.693 \\
 & Electrolyte/metabolic & 0.340 & 0.357 & 0.662 & 0.674 \\
 & Inflammatory & 0.276 & 0.289 & 0.679 & 0.696 \\
\hline
\end{tabular}
\end{table}

Table~\ref{tab:physionet_task_avg} reports endpoint-level results on PhysioNet 2012. Curve-NECA has higher mean PR-AUC and ROC-AUC than Raw across all four endpoints. The largest mean difference appears for neurological events, where PR-AUC changes from 0.461 to 0.557 and ROC-AUC from 0.689 to 0.766. The mean gains do not establish gains for every learner. The input mismatch described above also limits attribution to neighbor outcomes or event timing.

Table~\ref{tab:mimic_task_avg} reports the same aggregation on the MIMIC-IV subset. The largest mean difference again appears for neurological events. PR-AUC changes from 0.419 to 0.520, and ROC-AUC changes from 0.749 to 0.809. The other endpoints show smaller mean differences. The record-level split permits records from one subject in different partitions, so these results require subject-disjoint re-evaluation.

Overall, the reported experiments motivate further study of neighbor event curves. Additional controls are needed to separate the effect of event timing from retrieval inputs, measurement patterns, and the split procedure. The MIMIC-IV analysis covers 4,000 stays from a single data source, so it does not establish external validity.

\subsection{Neighbor Retrieval Ablation}

To examine retrieval sensitivity, we compare several early-window feature sets on neurological events. Table~\ref{tab:ablation_neighbor_retrieval} reports validation-set averages across learners and horizons. Retrieval with the 27 abnormality indicators has the highest values, with PR-AUC 0.550 and ROC-AUC 0.744. Retrieval with laboratory variables ranks second. The feature sets contain different variables, so this comparison cannot isolate the effect of the similarity measure.

\begin{table}[t]
\centering
\caption{Neighbor retrieval ablation on the neurological endpoint. Values are validation-set averages across learners and horizons.}
\label{tab:ablation_neighbor_retrieval}
\necatablestyle
\begin{adjustbox}{max width=\columnwidth}
\begin{tabular}{lcc}
\hline
\textbf{Retrieval strategy} & \textbf{Avg. PR-AUC} & \textbf{Avg. ROC-AUC} \\
\hline
Random neighbors & 0.446 & 0.666 \\
Demographic variables & 0.338 & 0.561 \\
Endpoint-specific variables & 0.385 & 0.615 \\
Physiological variables & 0.416 & 0.678 \\
Laboratory variables & 0.534 & 0.727 \\
Abnormality indicators & \textbf{0.550} & \textbf{0.744} \\
\hline
\end{tabular}
\end{adjustbox}
\end{table}

\subsection{Shuffled Event-Time Control}

The reported time-shuffle ablation assigns a randomly sampled finite event time to every retrieved neighbor, including neighbors without a recorded event. It then adjusts each curve column to approximately match its original mean. This procedure changes each neighbor's event status as well as its event time. Table~\ref{tab:time_shuffled_main} shows the resulting scores. The decrease cannot be attributed to timing alone. Testing that claim would require shuffling event times among neighbors with a recorded event by 42 hours while preserving their event status.

\begin{table}[t]
\centering
\caption{Full curve and the reported shuffled-curve control. Values are average PR-AUC across horizons. The difference is shuffled minus full. This control changes event status and event time.}
\label{tab:time_shuffled_main}
\necatablestyle
\begin{adjustbox}{max width=\columnwidth}
\begin{tabular}{llccc}
\hline
\textbf{Data} & \textbf{Endpoint} & \textbf{Full} & \textbf{Shuf.} & $\Delta$ \\
\hline
\multirow{4}{*}{PN12} & Neuro. & 0.519 & 0.355 & -0.164 \\
& Renal & 0.384 & 0.338 & -0.046 \\
& Electro./Metab. & 0.372 & 0.333 & -0.039 \\
& Inflam. & 0.536 & 0.489 & -0.047 \\
\hline
\multirow{4}{*}{MIMIC} & Neuro. & 0.493 & 0.342 & -0.151 \\
& Renal & 0.350 & 0.299 & -0.051 \\
& Electro./Metab. & 0.341 & 0.314 & -0.027 \\
& Inflam. & 0.286 & 0.258 & -0.028 \\
\hline
\end{tabular}
\end{adjustbox}
\end{table}

\subsection{Single-Horizon and Neighbor-Mean Ablations}
\label{sec:single_horizon}

The single-horizon variant adds only the neighbor event rate $z_{i,H}^{(g)}$ at the target horizon $H$. Full Curve-NECA adds all four rates in $\mathbf{z}_i^{(g)}$. Table~\ref{tab:single_vs_full} shows that the single-horizon rate already improves over Raw XGBoost. The full curve adds a smaller increment, especially for neurological events. The table's difference is single-horizon PR-AUC minus full-curve PR-AUC.

\begin{table}[t]
\centering
\caption{Single-horizon neighbor event rate versus full Curve-NECA on PhysioNet 2012 using XGBoost. Values are average PR-AUC across horizons.}
\label{tab:single_vs_full}
\necatablestyle
\begin{adjustbox}{max width=\columnwidth}
\begin{tabular}{lcccc}
\hline
\textbf{Endpoint} & \textbf{Raw} & \textbf{Full} & \textbf{Single} & $\Delta_{\text{S-F}}$ \\
\hline
Neuro. & 0.331 & 0.443 & 0.417 & -0.026 \\
Renal & 0.328 & 0.366 & 0.358 & -0.008 \\
Electro./Metab. & 0.326 & 0.341 & 0.331 & -0.010 \\
Inflam. & 0.490 & 0.513 & 0.507 & -0.006 \\
\hline
\end{tabular}
\end{adjustbox}
\end{table}

The neighbor-mean variant appends averages of retrieved neighbors' early-window variables without their event times. On neurological events, its validation PR-AUC is 0.509, compared with 0.443 for Raw and 0.550 for Curve-NECA. This comparison does not separate neighbor outcomes, cross-system inputs, and timing.

\section{Discussion and Conclusion}

NECA adds cumulative event rates from retrieved training records to an ICU predictor. The target-horizon neighbor rate explains much of the observed improvement, while the other curve points add smaller increments. Classifiers are fitted independently by horizon and may therefore produce decreasing risks across longer horizons. We have not measured the frequency of such predictions or assessed calibration.

The largest difference concerns neurological events, but we have not tested whether cross-system retrieval variables explain it. A low recorded GCS can reflect sedation, intubation, or measurement practice. Clinical interpretation requires a targeted sensitivity analysis.

The findings remain preliminary. Endpoint rules were not clinically adjudicated. Missing measurements or insufficient follow-up can be coded as no recorded event. The Raw tabular models lack some retrieval variables. MIMIC-IV uses a record-level split, training event curves were not cross-fitted, and $K=100$ was fixed without validation-based selection. The shuffle control changes event status, and the mean differences lack paired uncertainty estimates. These limitations prevent claims about event-time specificity, generalization, or clinical benefit.

\appendix

\section{Retrieval Indicator Definitions}
\label{app:flag_definitions}

The retrieval vector has 27 entries from the first six ICU hours. Thirteen basic indicators are age $\geq65$, hypotension (minimum SBP $<90$ or MAP $<65$), tachycardia (maximum HR $>110$ or last HR $>100$), hypoxemia (minimum SpO$_2<92$, or PaO$_2<60$ when SpO$_2$ is unavailable), acidosis (minimum pH $<7.35$ or bicarbonate $<22$), lactate $>2$, renal abnormality (maximum creatinine $>1.5$ or BUN $>30$), worsening creatinine (last $>1.1\times$ first $+0.05$), inflammation (maximum WBC $>12$ or temperature $>38^\circ$C), bilirubin $>2$, recorded ventilation, at least 60 measurements, and last GCS $\leq12$.

Six system indicators cover hemodynamic findings (hypotension or tachycardia), respiratory findings (hypoxemia or last RR $>24$), renal findings (abnormal or worsening creatinine), inflammation, hepatic findings, and coagulation (minimum platelets $<100$). Five interaction indicators capture shock-like findings (hypotension and tachycardia with acidosis or elevated lactate), cardio-respiratory findings (hypotension with hypoxemia or tachypnea), renal-perfusion findings (hypotension with renal abnormality or last urine output $<40$), inflammation with a renal, respiratory, or hemodynamic abnormality, and at least three abnormal systems.

The final three entries are abnormal-system count divided by five, severe-signal count divided by five, and ICU-type code divided by four. Severe signals are hypotension with tachycardia, hypoxemia with hypotension, renal abnormality with hypotension, acidosis or elevated lactate, and low GCS. Missing measurements make binary rules false, and missing ICU type is zero. Some retrieval variables are unavailable to the endpoint-specific Raw tabular predictor.

\FloatBarrier

\end{document}